\documentclass[final]{cvpr}

\usepackage{times}
\usepackage{epsfig}
\usepackage{graphicx}
\usepackage{amsmath}
\usepackage{amssymb}

\usepackage{bm}
\usepackage{multirow}
\usepackage{multicol}
\usepackage{floatrow}
\usepackage[normalem]{ulem}
\usepackage{graphics}
\usepackage{algpseudocode}
\usepackage{caption}
\usepackage{makecell}
\usepackage{booktabs}
\usepackage{scalerel}
\usepackage{paralist}
\usepackage[table]{xcolor}
\usepackage{color,colortbl}

\usepackage[pagebackref=true,breaklinks=true,colorlinks,bookmarks=false,citecolor=blue,linkcolor=blue]{hyperref}
\usepackage{xcolor}
\definecolor{gray}{rgb}{0.5,0.5,0.5} 
\definecolor{frenchblue}{rgb}{0.0, 0.45, 0.73}
\definecolor{gray}{rgb}{0.5,0.5,0.5} 
\definecolor{green}{rgb}{0, 0.4, 0} 
\definecolor{orange}{rgb}{1, 0.5, 0} 	
\definecolor{mahogany}{rgb}{0.75, 0.25, 0.0}
\definecolor{purple}{rgb}{0.6, 0, 0.6}
\definecolor{darkgreen}{rgb}{0, 0.4, 0.4} 
\definecolor{teal}{rgb}{0.0, 0.5, 0.5}
\definecolor{aaaa}{rgb}{0.55, 0.1, 0.7}
\definecolor{red}{rgb}{1.0, 0, 0}
\definecolor{plotpurple}{rgb}{0.2353, 0.2, 0.90196}
\definecolor{plotorange}{rgb}{1.0, 0.6, 0.2}
\definecolor{plotgreen}{rgb}{0.2, 0.784313, 0.2}
\definecolor{plotred}{rgb}{1.0, 0.2, 0.392}
\definecolor{lightgray}{gray}{0.9}
\definecolor{LightCyan}{rgb}{0.88,1,1}
\definecolor{baselinecolor}{gray}{.9}
\definecolor{plotmagenta}{rgb}{0.839, 0, 1}

\newcommand\Bstrut{\rule[-0.3ex]{0pt}{0pt}}
\newcommand\Tstrut{\rule{0pt}{2ex}}

\pdfoutput=1

\def\cvprPaperID{} 
\def\confName{ICCV}
\def\confYear{2023}

\begin{document}

\title{Delving into Motion-Aware Matching for Monocular 3D Object Tracking}

\author{Kuan-Chih Huang$^1$ \quad\quad Ming-Hsuan Yang$^{1,2,3}$ \quad\quad Yi-Hsuan Tsai$^2$ \vspace{0.2cm}\\
$^1$University of California, Merced \quad $^2$Google \quad $^3$Yonsei University
}

\maketitle

\pagestyle{empty}  
\thispagestyle{empty}

\begin{abstract}
    Recent advances of monocular 3D object detection facilitate the 3D multi-object tracking task based on low-cost camera sensors.
    In this paper, we find that the motion cue of objects along different time frames is critical in 3D multi-object tracking, which is less explored in existing monocular-based approaches.
    %
    To this end, we propose MoMA-M3T, a framework that mainly consists of three motion-aware components.
    First, we represent the possible movement of an object related to all object tracklets in the feature space as its motion features.
    Then, we further model the historical object tracklet along the time frame in a spatial-temporal perspective via a motion transformer.
    Finally, we propose a motion-aware matching module to associate historical object tracklets and current observations as final tracking results.
    We conduct extensive experiments on the nuScenes and KITTI datasets to demonstrate that our MoMA-M3T achieves competitive performance against state-of-the-art methods. 
    Moreover, the proposed tracker is flexible and can be easily plugged into existing image-based 3D object detectors without re-training.
    Code and models are available at \url{https://github.com/kuanchihhuang/MoMA-M3T}.
\end{abstract}

\section{Introduction}

\begin{figure}
    \centering
\includegraphics[width=0.98\linewidth]%
{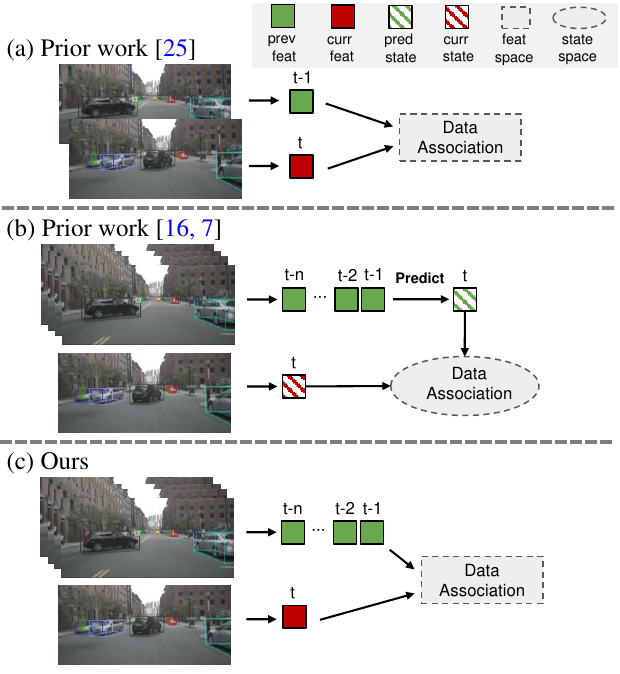}
    \caption{
    \textbf{
     Comparisons of different association methods in monocular 3D object tracking.
    } 
    (a) Time3D~\cite{li2022time3d} learns to match 3D object features in adjacent frames.
    (b) QD-3DT~\cite{hu2022monocular} and DEFT~\cite{Chaabane2021deft} utilize the object's previous features to predict their current states, and match with the observations in the output space.
    (c) Our approach directly aggregates the object's previous features and matches them with current observations in the feature space.
    }
    \label{fig:example}
    \vspace{-6pt}
\end{figure}

3D Multi-Object Tracking (3D MOT) is a crucial problem for various applications like autonomous driving. 
Numerous LiDAR-based methods \cite{Weng2020_AB3DMOT, weng2020gnn3d} have achieved remarkable results thanks to powerful 3D object detectors \cite{Lang2019pointpillars, shi2019pointrcnn, he2020sassd, shi2020pv}. 
Due to the lower cost of camera sensors, some image-based 3D object detection approaches \cite{Mousavian2017deep3dbox,ku2019monopsr, ma2019am3d, brazil2020kinematic, huang2021bevdet, li2022bevformer} receive much attention and achieve promising performance, and thus enable 3D object tracking based on merely the camera.

One straightforward approach to deal with monocular 3D object tracking is to match object features in adjacent frames \cite{zhou2020tracking, li2022time3d} (see Figure \ref{fig:example}(a)).
Although significant progress has been made, these methods may still fail to capture multi-frame motion information of objects. 
To tackle the long-range dependency, another line of work \cite{hu2022monocular, Chaabane2021deft} predicts the object states from the historical observations, in which the predicted and observed states in the current frame are in the output space that explicitly contains the object information, \eg, location and pose of the object (see Figure \ref{fig:example}(b)).
However, these approaches may suffer from noisy observations of object states predicted by the inaccurate monocular 3D object detector.

For the above-mentioned methods, one critical step is data association, in which the goal is to match observations across historical time frames and produce final tracking results.
Therefore, in monocular 3D MOT, two main challenges are 1) how to obtain the long-range observations that can provide richer information for data association? 2) what are the better representations the algorithm utilizes as observations, in order to mitigate the problem of matching under noisy observations from the inaccurate monocular 3D detector?
In this paper, we propose MoMA-M3T, a motion-aware matching approach for monocular 3D MOT, to handle these two challenges (see Figure \ref{fig:example}(c)). Our main idea is to encode the multi-frame motion information of the object tracklets, i.e., their historically \textit{relative} positions, into a \textit{feature} space for data association, instead of encoding their \textit{absolute} locations in the \textit{output} space. %
To this end, the object movements encoded in the learned representations can be used for matching between tracklets and current object observations.

Specifically, MoMA-M3T consists of three main components: 1) we first use a motion encoder to encode the 3D object information, \eg, relative position and object size/heading angle, into a motion-aware feature space; 2) Then, these encoded features also form a motion feature bank to record historical features, followed by a motion transformer module to generate spatial-temporal motion features as representations of object tracklets; 3) Finally, a motion-aware matching module to generate tracking results is introduced for data association between object observations and tracklets based on motion features.
Moreover, our method that considers motion features enables the feasibility of applying learning strategies. We adopt a contrastive learning objective that samples several subsets of different object trajectories and learns better feature representations, \eg, data points from the same trajectory but in augmented views are positive samples.

Extensive experiments on nuScenes~\cite{nuscenes2020caesar} and KITTI~\cite{Geiger2012kitti} datasets demonstrate that our method achieves state-of-the-art performance based on monocular camera sensors.
In addition, we show the benefit of our proposed components, including the usage of motion features, motion transformers, and motion-aware matching. More interestingly, we present the robustness of MoMA-M3T by plugging our learned modules with frozen weights into the same framework, but based on detection outputs from different 3D object detectors. Results show that our motion-aware approach generalizes well to various pre-trained detectors.

The main contributions of this work are as follows:

\begin{itemize}
    \item We present MoMA-M3T, a framework that introduces motion features with a motion-aware matching mechanism for monocular 3D MOT.
    
    \item We propose a motion transformer module that captures the movement of object tracklets in a spatial-temporal perspective, enabling robust motion feature learning. %
    
    \item Extensive experiments on nuScenes and KITTI datasets show that our method achieves competitive performance based on monocular sensors, with the flexibility to apply various pre-trained 3D detectors.
\end{itemize}

\section{Related Work}

\noindent {\bf Monocular 3D Object Detection.}
Image-based 3D object detection has gained much attention recently due to the low-cost camera sensors.
Numerous approaches \cite{brazil2019m3drpn, Ma2021monodle, kumar2021groomed,andrea2019monodis, liu2021ground,Zhang2021MonoFlex,wang2021fcos3d, wang2021PGD, epropnp} perceive 3D objects on the image plane by relying on geometric relationships \cite{Simonelli2020MoVi3D, Shi2021MonoRCNN}, such as object size \cite{Shi2021MonoRCNN}, keypoints \cite{liu2020SMOKE, li2020rtm3d}, or depth uncertainty \cite{lu2021gupnet, wang2021PGD}.
To improve the 3D reasoning ability, several approaches \cite{ding2020d4lcn, huang2022monodtr,Chen_2022_stereo, wang2019pseudo,Zou2021dfr} leverage depth information to facilitate object detection.
In addition, some works \cite{yue2021detr3d, huang2021bevdet, li2022bevformer, liu2022petr} focuses on developing multi-camera 3D object detection systems. 
These methods learn the bird-eye view representations of the surrounding scenarios by fusing information from multiple cameras.
Instead of designing a powerful monocular 3D detector, our work targets on establishing a robust motion tracker to associate noisy monocular 3D observations.

\smallskip \noindent {\bf Multi-Object Tracking.}
With the rapid advances of object detection, Multi-object tracking (MOT) \cite{zhou2020tracking, zhang2022bytetrack,Wu2021TraDeS, tokmakov2021learning,zhou2022global} has been extensively explored in the 2D image space. 
Most state-of-the-art approaches adopt the tracking-by-detection paradigm \cite{Bewley2016_sort,Wojke2017deepsort}, which detects the objects first, followed by the tracking module that leverages different information such as visual appearances \cite{zhang2021fairmot, Chaabane2021deft} or motion cues \cite{Wojke2017deepsort, zhang2022bytetrack}, to associate the object boxes. 

Extending from 2D MOT techniques, existing 3D MOT approaches mainly rely on the high-quality LiDAR detector to track objects in 3D space. 
AB3DMOT \cite{Weng2020_AB3DMOT} adopts a 3D IOU similarity metric, and the Kalman filter \cite{kalman1960kalman} to predict and update the state of objects. 
CenterPoint\cite{yin2021center} adds a learnable velocity estimation head to replace the Kalman filter to perform tracking, while GNN3DMOT \cite{weng2020gnn3d} and PTP \cite{Weng2021_PTP} exploit graph neural networks to integrate the appearance and motion features from LiDAR and image information. 
Furthermore, to avoid handicraft or heuristic matching steps in the previous pipeline, SimTrack \cite{Luo2021simtrack} introduces an end-to-end joint detection and tracking model to associate data implicitly.
In this paper, different from relying on LiDAR signals as the above-mentioned methods, we develop the 3D MOT approach based on purely monocular cameras.

\begin{figure*}[t]
\centering

\includegraphics[width=.98\textwidth]{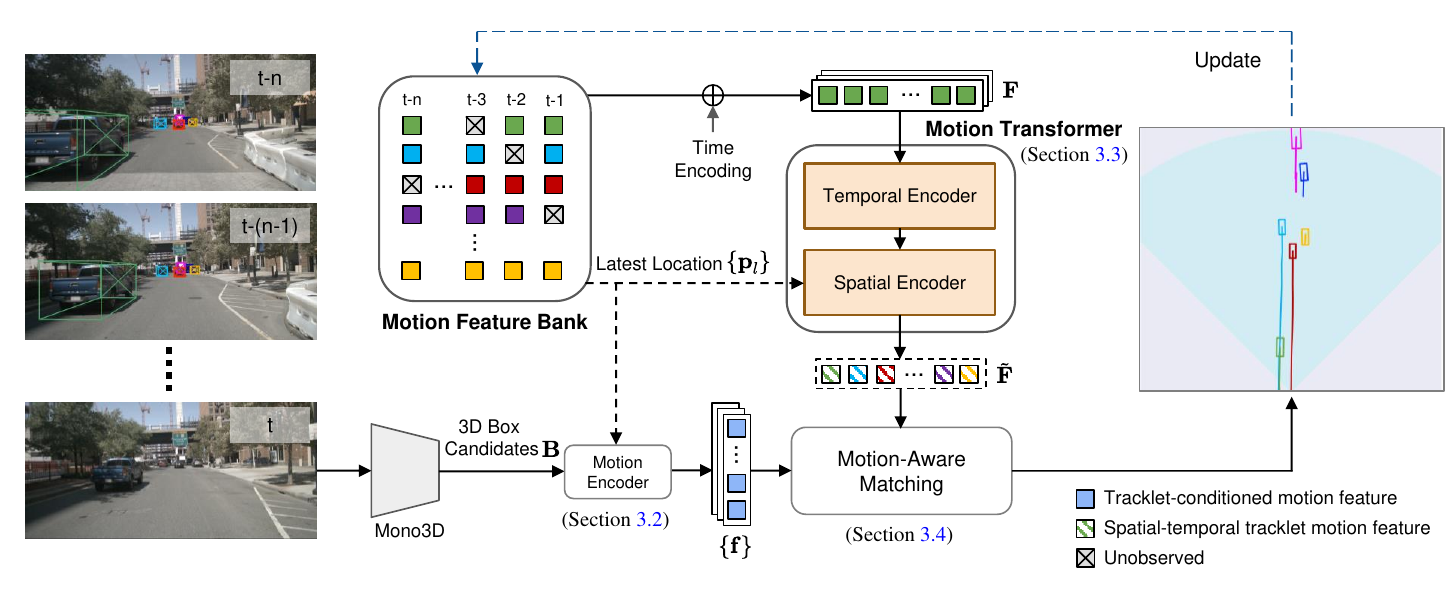} %
\caption{\textbf{Overall framework of the proposed MoMA-M3T.} 
At each timestamp, we leverage a monocular 3D object detector to generate 3D bounding box candidates $\mathbf{B}$.
Then, we take the latest 3D positions $\{\mathbf{p}_l\}$ of tracklets to generate all possible movements for each tracklet-detection pair, followed by a motion encoder to extract the tracklet-conditioned motion features $\{\mathbf{f}\}$ (Section \ref{sec:motion_repre}). 
On the other hand, a motion transformer module is applied to aggregate the motion cues $\mathbf{F}$  temporally and spatially from different timestamps, resulting in motion features $\tilde{\mathbf{F}}$ for each tracked object (Section \ref{sec:mtr}). Finally, a motion-aware matching strategy is adopted to associate the learned motion features between tracklets and detections (Section \ref{sec:mm}).
}
\label{fig:arch}
\vspace{-13pt}
\end{figure*}

\smallskip \noindent {\bf Monocular 3D Multi-Object Tracking.} 
Compared with LiDAR-based 3D MOT, camera-based 3D MOT \cite{Marinello2022triplet, zhou2020tracking, cc3dt} is a challenging task due to the inaccurate object depth estimation. 
Early methods \cite{zhou2019objects, tokmakov2021learning, Wu2021TraDeS} mostly extend from 2D MOT frameworks to track 3D objects in the image plane, which may achieve undesirable performance since it cannot well capture the motion of objects. 
Furthermore, QD-3DT \cite{hu2022monocular} and DEFT \cite{Chaabane2021deft} jointly learn the objects' state and their Re-ID features, followed by an LSTM-based module for modeling the movement of objects. 
Recently, inspired by the success of transformer-based 2D MOT framework MOTR \cite{zeng2021motr}, MUTR3D \cite{zhang2022mutr3d} leverages 3D track queries to associate objects based on the multi-camera detector\cite{yue2021detr3d}. 
On the other hand, Time3D \cite{li2022time3d} jointly learns the 3D detection and tracking from a monocular stream in an end-to-end manner, which utilizes the transformer to model the relationship between objects within adjacent frames.

However, less effort has been made to tackle one important problem of monocular 3D MOT, \ie, matching inaccurate and noisy predictions in multi-frame
observations.
Thus, we focus on modeling object tracklets and detections with motion representations, while designing motion-aware modules to help the learning process, \eg, motion transformer and motion-aware matching in the feature space.

\section{Proposed Approach}
\subsection{Framework Overview}
\label{sec:loss}
Given the detected 3D bounding box candidates ${\mathbf{B}}_t=\{\mathbf{b}_{t}\}$ at frame $t$ from the monocular 3D object detectors \cite{wang2021PGD, epropnp, wang2021fcos3d}, where $\mathbf{b}=(\mathbf{p},\theta,h,w,l)$ denotes an object's 3D position $\mathbf{p}=(x,y,z)$, heading angle $\theta$, and object size $(h,w,l)$, we aim to perform online 3D MOT to find a set of tracklets ${\mathbf{T}}_t=\{\mathbf{\tau}_{t}\}$.
In this paper, we propose a motion-aware matching approach, MoMA-M3T, for monocular 3D MOT, following the tracking-by-detection paradigm to associate observations and object tracklets.

As shown in Figure \ref{fig:arch},
MoMA-M3T mainly consists of three modules: the motion encoder, the motion transformer, and the motion-aware matching module. At each timestamp $t$, we first utilize an encoder to generate possible motion-aware feature candidates based on the movement between observations and tracklets (Section \ref{sec:motion_repre}). Then a motion transformer is exploited to aggregate motion representations of tracklets across different time frames in a spatial-temporal perspective (Section \ref{sec:mtr}). 
Consequently, we learn the affinity matrix for identity matching based on the motion-aware representations of observations and tracklets with the motion-aware matching module (Section \ref{sec:mm}).

\subsection{Motion Feature Generation}

\label{sec:motion_repre}
Unlike the previous work \cite{li2022time3d} that directly encodes the absolute positions for objects, we express them with motion representations based on their movement vectors along different time frames, 
which facilitates matching under inaccurate observations from the monocular 3D detector.

\smallskip \noindent {\bf Motion Representation.}
Consider an object's two global positions $\mathbf{p}_a, \mathbf{p}_b \in \mathbb{R}^{3}$, the relative movement from $b$ to $a$ can be expressed as:
\begin{equation}
    \mathbf{r}_{a|b}=\mathbf{p}_a-\mathbf{p}_b.
    \label{eq:relative}
\end{equation}
In addition, we define the motion state of an object at timestamp $t$ as $\textbf{s}_t = (\mathbf{r}, \theta, h, w, l)_t$ with its heading angle and size, where $\mathbf{r}$ indicates the position movement from the previous frame to the current frame.
To obtain motion features of the object, we apply a motion encoder via a multi-layer perceptron (MLP) to describe state information:
\begin{equation}
    \mathbf{f}_t = \rm{MLP}(\textbf{s}_t) \in \mathbb{R}^{C},
    \label{eq:motionfeat}
\end{equation}
where $C$ is the feature dimension. As such, $\mathbf{f}_t$ can be used to express the motion features of any object at frame $t$. 

\smallskip \noindent {\bf Tracklet-conditioned Motion Feature.}
For a single frame observation, since its previous location is undetermined before the tracking association process, we take the latest positions of all tracklets as their last locations to generate all possible motion features.

\begin{figure}[!t]
    \centering
    \includegraphics[width=0.95\linewidth]{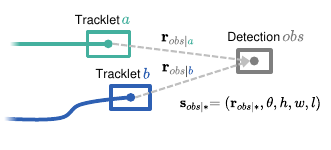}
    \vspace{-5mm}
    \caption{
    \textbf{
    Tracklet-conditioned motion state.
    } 
    For any observation with estimated 3D position $\mathbf{p}_{obs}$, we calculate the relative movement $\mathbf{r}_{obs|*}$ to the latest position of any active tracklet, \eg, $a$ and $b$. With other detected object information, \ie, heading angle $\theta$ and object size ($h,w,l$), we generate the object's tracklet-conditioned motion state $\mathbf{s}_{obs|*}$.
    }
    \label{fig:motion}
    \vspace{-3mm}
\end{figure}

Specifically, consider $M$ tracklets with their latest positions $\mathbf{P}_l=\{\mathbf{p}_{l}\}$
and $N$ observations in the current frame  with the estimated positions $\mathbf{P}_{obs}=\{\mathbf{p}_{obs}\}$, we can adopt \eqref{eq:relative} to calculate all possible movements between detections and tracklets as $\mathbf{r}_{\{obs|l\}} \in \mathbb{R}^{N \times M \times 3}$. 
Note that, if there is no tracklet, we set the relative movements of objects as zero.
Next, we generate all candidate motion states and utilize \eqref{eq:motionfeat} to extract all candidate motion features for the current observations, which are referred to as tracklet-conditioned motion features. We utilize an example shown in Figure \ref{fig:motion} to illustrate the process of motion state generation. 

\smallskip \noindent {\bf Motion Feature Bank.}
We create a motion feature bank to maintain historical motion features $\textbf{F}_{bank} \in \mathbb{R}^{N_{max} \times T_{max} \times C}$ and global 3D positions $\textbf{P}_{bank} \in \mathbb{R}^{N_{max} \times T_{max} \times 3}$ for all tracklets, where $N_{max}$ is the maximum number of tracked objects, and $T_{max}$ denotes the maximum time length.
After the tracking association process in each time frame (details introduced in the later section), we store the tracked objects' latest positions and their tracklet-conditioned motion features in the feature bank.

\subsection{Motion Transformer}
\label{sec:mtr}
To capture the motion behavior of different tracked objects, inspired by the transformer's success in modeling sequential data, we propose a motion transformer to express the object's motion representations from a spatial-temporal perspective, which consists of three modules: time encoding, temporal encoder, and spatial encoder.

\smallskip \noindent {\bf Input and Time Encoding.}
For the input of the transformer, we take the latest $T$-frame features of each tracklet from the motion feature bank. 
Considering the objects may be non-consecutive in certain frames due to the occlusion or undetected results (denoted as grey grids in the motion feature bank of Figure \ref{fig:arch}), we add a learnable time positional embedding to make the model aware of temporal cues. Specifically, we take the time differences between the historical and current frames, and then apply a learnable positional encoding to learn the temporal cues.

\smallskip \noindent {\bf Temporal Encoder.}
To extract the temporal information for each tracklet, we exploit a transformer 
as the temporal encoder to model the object's motion representations along the temporal dimension. 
Specifically, considering the input motion feature $\mathbf{F} \in \mathbb{R}^{T \times C}$ of any tracklet from the motion feature bank $\textbf{F}_{bank}$
along $T$ frames,
we prepend a learnable motion token $\mathbf{F}_{m} \in \mathbb{R}^{1 \times C}$ to the sequence following BERT \cite{devlin2018bert}. Then, the concatenated features are fed to the multi-head self-attention encoder layers:
\begin{align}
  &  \mathbf{Q} = [\mathbf{F},\mathbf{F}_m]\mathbf{W}_q, \mathbf{K} = [\mathbf{F},\mathbf{F}_m]\mathbf{W}_k, \mathbf{V} = [\mathbf{F},\mathbf{F}_m]\mathbf{W}_v, \nonumber \\
  &  [\hat{\mathbf{F}},\hat{\mathbf{F}}_m]={\rm FFN}({\rm{MultiHead}}(\mathbf{Q},\mathbf{K},\mathbf{V})),
\label{eq-att}
\end{align}
where $\mathbf{W}_*$ are learnable parameters for the temporal encoder, and $[\cdot,\cdot]$ means the feature concatenation operation. 
We use one linear layer followed by the ReLU activation to build our feed-forward network $\rm{FFN}$ (see  \cite{vaswani2017SA} for details about self-attention layer $\rm{MultiHead}$). 
Consequently, we output the learned motion token $\mathbf{\hat F}_{m} \in \mathbb{R}^{1 \times C}$ to reflect motion representations of the tracked object, which is then sent to the subsequent spatial encoder module.

\smallskip \noindent {\bf Spatial Encoder.}
We observe that the states of objects (\eg, locations) may depend on other objects in the same scene. 
Thus, we exploit a spatial encoder after the temporal module to capture spatial dependencies among tracklets, including tracklets’ states and their relationships.

Considering the aggregated motion features from the temporal encoder only encode the local relative movement, we further introduce the absolute position of objects as global information.
For $M$ tracklets with their features $\{\mathbf{\hat F}_{m}\} \in \mathbb{R}^{M \times C}$ via \eqref{eq-att} and latest locations $\{\mathbf{p}_l\} \in \mathbb{R}^{M \times 3}$ from $\textbf{P}_{bank}$, 
we use two linear layers to encode the global positional features $\mathbf{X}_p=\rm{MLP}(\{\mathbf{p}_l\})$.
Then, we incorporate the position and motion features into the transformer: 
\begin{align}
 & \mathbf{Q}^s \rm{=} \{\mathbf{\hat F}_{m}\}\mathbf{W}^s_q,  
 \mathbf{K}^s  \rm{=} [\mathbf{X}_p,\{\mathbf{\hat F}_{m}\}]\mathbf{W}^s_k, 
 \mathbf{V}^s \rm{=} [\mathbf{X}_p,\{\mathbf{\hat F}_{m}\}]\mathbf{W}^s_v,  \nonumber \\
  &  \tilde{\mathbf{F}}  ={\rm FFN}({\rm{MultiHead}}(\mathbf{Q}^s,\mathbf{K}^s,\mathbf{V}^s)),
\label{eq-att-s}
\vspace{-2mm}
\end{align}
where $\mathbf{W}^s_*$ are learnable parameters for the spatial encoder.
Finally, we output the spatially interacted features $\tilde{\mathbf{F}} \in \mathbb{R}^{M \times C}$ for tracklets to represent their final motion features, which can be used for the matching process.

\begin{figure}[!t]
    \centering
    \includegraphics[width=\columnwidth]{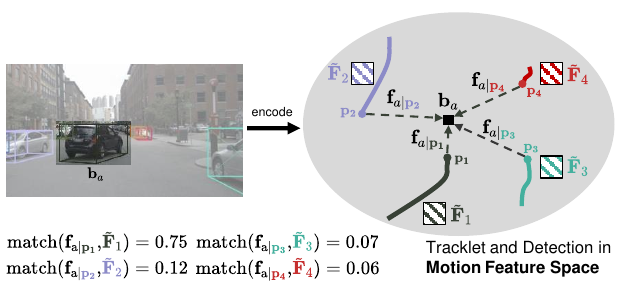} %
    \vspace{-7mm}
    \caption{\textbf{Motion-aware matching.} 
    For $M$ tracklets with their motion features $\tilde{\mathbf{F}}$ and latest locations $\mathbf{p}$ ($M=4$ for illustration), given any observation $\mathbf{b}_a$, we generate tracklet-conditioned motion features $\mathbf{f}_{a|{\mathbf{p}}}$ based on the representations described in Section \ref{sec:motion_repre}. We use an $\rm MLP$ layer to predict a pairwise matching score, as the difference between detection’s and tracklet’s motion features.
    }
    \label{fig:mm}
\end{figure}

\subsection{Motion-Aware Matching Learning}
\label{sec:mm}
After obtaining the motion features $\tilde{\mathbf{F}}$ for tracklets and the tracklet-conditioned motion features $\{\mathbf{f}\}$ for observations in the current frame, we aim to perform affinity learning to solve data association for detection-tracklet pairs.

\smallskip\noindent {\bf Matching in Motion Feature Space.}
In Figure \ref{fig:mm}, consider $M$ tracklets with their motion features $\tilde{\mathbf{F}} \in \mathbb{R}^{M \times C}$ and $N$ detected objects $\{\mathbf{b}_i\}$ with their tracklet-conditioned motion features $\{\mathbf{f}_{i|{\mathbf{p}}}\} \in \mathbb{R}^{N \times M \times C}$ (based on the tracked objects' latest positions ${\mathbf{p}}$ as described in Section \ref{sec:motion_repre}), our goal is to output a matching score between 0 and 1 to indicate whether any detection-tracklet pair has the same identity. 
We use an $\rm MLP$ layer to learn a mapping function with the input of the difference between detection's and tracklet's motion features, followed by a sigmoid function:
\begin{equation}
    \textbf{A}_{ij} = \rm Sigmoid(\rm MLP(\mathbf{f}_{i|{\mathbf{p}_j}} - \tilde{\mathbf{F}}_j)),
    \label{eq:affinity}
\end{equation}
where $\textbf{A}_{ij}$ is the probability of the detection $i$ and the tracklet $j$ belonging to the same identity.
We apply a binary focal loss $\rm{FL}$ \cite{lin2018focal} to learn the matching process:
\begin{align}
    \mathcal{L}_{match}=\frac{1}{N \cdot M}\sum_{i}^{N}\sum_{j}^{M}
    \rm{FL}(\mathbf{A}_{ij}, \mathbf{\hat A}_{ij}),
\end{align}
where $\mathbf{\hat A}$ is the ground truth affinity value, \ie, 1 or 0, depending on whether the pairs are the same object or not.

\begin{figure}[!t]
    \centering
    \includegraphics[width=\columnwidth]{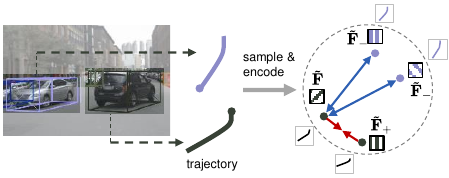} 
    \vspace{-6mm}
    \caption{\textbf{Contrastive motion learning strategy.} We randomly sample the subsets of the trajectory for all objects and apply contrastive learning to construct a robust feature space, encouraging the motion features from the same trajectory to be similar ($\tilde{\mathbf{F}}$ and $\tilde{\mathbf{F}}_+$), and dissimilar to different trajectories ($\tilde{\mathbf{F}}$ and $\tilde{\mathbf{F}}_-$).  
    }
    \label{fig:contras}
\end{figure}

\smallskip\noindent {\bf Contrastive Motion Feature Learning.}
Due to the occlusion and the inaccurate predictions, monocular 3D object detection results are generally noisy, which is challenging for the model to learn the motion pattern of the objects. 
To alleviate this, we propose a contrastive motion learning strategy that learns robust motion representations for each tracklet as illustrated in Figure \ref{fig:contras}.

Considering all tracked objects in a video, we randomly sample $k$ subset of their trajectories (positions along different timestamps). 
Based on contrastive learning, the trajectory subsets sampled from the same tracklet should have similar motion representations, and the distinct trajectory subsets should have dissimilar representations.
For all sampled trajectories, we can utilize the motion transformer described in Section \ref{sec:mtr} to encode their motion feature $\tilde{\mathbf{F}}$ 
and apply a contrastive loss \cite{khosla2020supervised} 
for representation learning:
\begin{equation}
    \mathcal{L}_{\text{con}}
    = -\frac{1}{|\mathbf{N}_p|}\sum_{(i,j) \in \mathbf{N}_p} \log \frac{
        \exp (\tilde{\mathbf{F}}_i \cdot \tilde{\mathbf{F}}_j / \tau)
    }{
        \sum_{(i,k) \in \mathbf{N}_a} { \exp (\tilde{\mathbf{F}}_i \cdot \tilde{\mathbf{F}}_k / \tau)},
        }    
\end{equation}
where $\tau$ is the temperature parameter, which is set to 0.1 in our implementation. ${\mathbf{N}_p}$ is the set of positive pairs sampled from the same trajectory, while negative pairs are in the set of ${\mathbf{N}_a} \backslash {\mathbf{N}_p}$, in which ${\mathbf{N}_a}$ contains all the samples.
The process encourages the motion embeddings $\tilde{\mathbf{F}}_i$ and $\tilde{\mathbf{F}}_j$ from the same tracklet to be similar, while $\tilde{\mathbf{F}}_i$ and $\tilde{\mathbf{F}}_k$ from the different ones should be dissimilar.

\smallskip\noindent {\bf Overall Objectives.}
The overall training loss of our network is defined as the summation of the matching loss and the contrastive loss: $\mathcal{L}=\mathcal{L}_{match}+\mathcal{L}_{con}$.

\begin{table*}[!t]
    \centering
    \footnotesize
    \setlength{\tabcolsep}{4.5pt}
    \begin{tabular}{llccccccc}  
        \toprule
        Method & Reference & AMOTA(\%)$\uparrow$  & AMOTP(m)$\downarrow$ & MOTA(\%)$\uparrow$  & MOTP(m)$\downarrow$ & MOTAR(\%)$\uparrow$  & MT$\uparrow$ & ML$\downarrow$  \\ 
        \midrule
        CenterTrack \cite{zhou2020tracking}     & ECCV'20     & 4.6 & 1.543  & 4.3& 0.753 & 23.1  & 573 & 5235 \\ 
        TraDeS \cite{Wu2021TraDeS}$\dagger$     & CVPR'21     & 5.9 & 1.49     & -    & -     & -    & - & - \\ 
        PermaTrack \cite{tokmakov2021learning}  & ICCV'21     & 6.6 & 1.491  & 6.0& 0.724 & 32.1  & 652 & 5065 \\
        DEFT \cite{Chaabane2021deft}            & CVPRw'21    & 17.7 & 1.564 & 15.6& 0.770 & 48.4  & 1951 & 3232 \\
        QD-3DT \cite{hu2022monocular}           & PAMI'22     & 21.7 & 1.550  & 19.8& 0.773 & 56.3  & 1893 & 2970 \\
        Time3D \cite{li2022time3d}$\dagger$     & CVPR'22     & 21.4 & \textbf{1.36}  
         & 17.3& 0.75 & -    & 
        - & - \\
        \midrule
        MoMA-M3T (Ours)                           & - & 24.2 & 1.479  & 21.3 & 0.713& 58.1 & 1968 & 3026 \\ 
        MoMA-M3T (Ours)$\ddagger$                           & - & \textbf{28.5} & 1.416  & \textbf{24.6} & \textbf{0.695}& \textbf{62.3}  & \textbf{2236} & \textbf{2642} \\ 
        \bottomrule
    \end{tabular}
    \vspace{-2mm}
    \caption{\textbf{3D MOT performance on the nuScenes test set for the single-camera tracking setting}.
    The best results are highlighted in $\textbf{bold}$.
    $\dagger$ indicates the results reported in their papers. 
    $\ddagger$ denotes using the detector \cite{wang2021PGD} trained with a longer schedule and data augmentations.
    }
    \label{tab:nuscnces_test}
\end{table*}

\subsection{Online Inference and Feature Update}
At each time frame $t$, after obtaining the 3D bounding box candidates from the monocular 3D object detector \cite{wang2021PGD, renNIPS15fasterrcnn, wang2021fcos3d, epropnp}, our modules first generate the motion features of tracklets through the motion transformer (Section~\ref{sec:mtr}), and tracklet-conditioned motion features of current object detections (Section~\ref{sec:motion_repre}).
Using the affinity matrix for each detection-tracklet pair via \eqref{eq:affinity}, the Hungarian algorithm \cite{Hungarian} is applied to match one-to-one pairs. 
If the matching score is larger than a threshold (\ie, 0.5 in this paper), this pair is selected as the tracking result. The motion features of the matched detection are also updated in the motion feature bank, along with their global positions.
Furthermore, we use the track rebirth strategy \cite{tracktor_2019_ICCV,zhou2020tracking} to retain unmatched tracklets until they are unmatched for 10 consecutive frames for handling the occlusion issue.

\section{Experiments}
\subsection{Experimental Setups}
\label{sec:setup}
\noindent{\bf {Datasets.}}
We evaluate our approach on nuScenes 3D MOT~\cite{nuscenes2020caesar} and KITTI 3D MOT~\cite{Geiger2012kitti}.
The nuScenes dataset contains 1000 real-world videos captured from six surrounding cameras with 7 object categories for the tracking task.
The dataset is officially split into 700, 150, and 150 sequences for training, validation, and testing. 
We follow \cite{hu2022monocular} to train our network on the keyframes and test on the $\it{full\ frames}$ for monocular 3D object tracking, which has higher frame rates.
The KITTI tracking dataset consists of 21 training and 29 testing scenes. As there is no official benchmark for the 3D tracking task on the KITTI dataset, we apply the metrics proposed in AB3DMOT\cite{Weng2020_AB3DMOT} for evaluation. 
We follow \cite{hu2022monocular} to divide the entire training set into a train set (13 scenes) and a validation set (8 scenes).

\smallskip\noindent{\bf Evaluation Metrics.}
On the nuScenes dataset, we utilize the official benchmark protocol to report the average performance for all categories, including AMOTA, AMOTP, MOTA, MOTP, MOTAR, mostly tracked (MT), and mostly lost (ML).
For the KITTI tracking dataset, we report the sAMOTA and AMOTA metrics \cite{Weng2020_AB3DMOT} of the car category for 3D evaluation.
We refer the readers to the supplementary material for more details.

\smallskip\noindent{\bf Implementation Details.}
Our approach is implemented in Pytorch on an NVIDIA 3090 GPU.
For training the proposed MoMA-M3T, we utilize the Adam optimizer for 100 epochs with batch size 128. 
The learning rate starts at 0.0001 and decays with a step of 0.5 decay rate every 20 epochs.
In each mini-batch, we randomly sample 16 tracklets with $T=6$ frames and 16 detections
for training the identity association process. In addition, we randomly sample $k=2$ subsets of each trajectory for motion contrastive learning, which results in 1 positive and $(16-1) \times k = 30$ negative samples for each trajectory.
For the motion feature bank, we set $N_{max}=50$ and $T_{max}=10$ with the channel number $C=128$. 
For nuScenes, we use PGD3D \cite{wang2021PGD} 
as our main monocular 3D detector. For KITTI, we utilize MonoDLE\cite{Ma2021monodle} for fair comparisons with existing methods. 
We include more details in the supplementary material.

\subsection{Main Results}

\vspace{-2mm}
\smallskip\noindent{\bf Monocular 3D MOT on nuScenes.}
To evaluate the 3D tracking performance based on monocular sensors on the nuScenes dataset, we follow \cite{zhou2020tracking, li2022time3d, hu2022monocular, Chaabane2021deft} to consider tracking and recognizing 3D objects from different cameras independently, which refers to as the single-camera tracking setting. 
In Table \ref{tab:nuscnces_test}, we report the tracking performance averaged of all categories on the nuScenes test set. 
Compared with other monocular 3D MOT methods, our approach achieves state-of-the-art results in most metrics. 
Specifically, compared to Time3D\cite{li2022time3d} using the detector with similar detection performance, i.e., 31.2 mAP (Time3D) $vs.$ 30.1 mAP (ours) on the nuScenes test set,
our tracking method performs better than Time3D by +2.8 AMOTA on average, which is the major metric in the benchmark.

\setlength{\tabcolsep}{0.035\linewidth}{
\begin{table}[t]
    \centering
    \footnotesize
    \begin{tabular}{lccc} 
    \toprule
        Method & Input &  sAMOTA$\uparrow$  & AMOTA$\uparrow$   \\ 
        \midrule
        QD-3DT \cite{hu2022monocular} & Mono  &  39.92     & 11.86  \\  %
        CenterTrack \cite{zhou2020tracking} & Mono & 42.28 & 11.37   \\
        MonoDLE\cite{Ma2021monodle}* & Mono &  46.16 & 13.00  \\
        \rowcolor{LightCyan}
        MoMA-M3T (Ours)            & Mono &\textbf{47.17} & \textbf{16.12} \\ 
        \bottomrule
        
    \end{tabular}
    \vspace{-2mm}
    \caption{\textbf{3D MOT performance on the KITTI validation set for the Car category at 0.25 IoU threshold} with the evaluation metric proposed in \cite{Weng2020_AB3DMOT}. * indicates using AB3DMOT as the tracker. All results are reproduced by ourselves based on their official codes and trained on the same data. We utilize the same detector as MonoDLE for fair comparisons.
    }
    \label{tab:kitti_val_car}
\end{table}
} %
\smallskip\noindent{\bf Monocular 3D MOT on KITTI.}
In Table \ref{tab:kitti_val_car}, we report the 3D MOT performance for the car category on the KITTI tracking dataset with the evaluation metric proposed in \cite{Weng2020_AB3DMOT} compared with different monocular-based methods, including QD3DT~\cite{hu2022monocular}, CenterTrack~\cite{zhou2020tracking}, and MonoDLE~\cite{Ma2021monodle} with the AB3DMOT tracker~\cite{Weng2020_AB3DMOT}. 
All baselines are reproduced by ourselves based on the official source codes
and trained under the same settings.
Overall, our approach achieves favorable performance against several monocular-based methods.
Specifically, compared to MonoDLE with the AB3DMOT tracker, our MoMA-M3T with the same detector obtains an improvement of +1.01 in sAMOTA and +3.12 in AMOTA, which validates the effectiveness of our motion tracker.

\smallskip{\noindent{\bf Runtime Speed.}}
We measure the inference speed of our motion tracker on a single NVIDIA 3090 GPU for processing the nuScenes validation set with batch size 1. Our tracker runs at 33.3 FPS on average.

\setlength{\tabcolsep}{0.017\linewidth}{
\begin{table}[t]
    \footnotesize
    \centering
    \begin{tabular}{l @{\hspace{4pt}} c @{\hspace{6pt}} c @{\hspace{8pt}} c @{\hspace{6pt}} c}
        \toprule
        & Representation & Matching Space & AMOTA$\uparrow$ & AMOTP$\downarrow$    \\ 
        \midrule
        (a) & Global &  Output  & 27.8 & 1.498     \\ %
        (b) & Global &  Feature  & 28.8 & 1.460   \\  \midrule%
        (c) & Motion &  Output  & 28.7 & 1.470       \\ %
        \rowcolor{LightCyan}
        (d) & Motion &  Feature  & 30.7 & 1.436      \\ %
         \rowcolor{LightCyan}
        (e) &Motion &  Feature$\dagger$  & \textbf{31.1} & \textbf{1.432}   \\
        \bottomrule
    \end{tabular}
    \vspace{-2mm}
    \caption{\textbf{Analysis of the importance of different representations and matching space} on the nuScenes validation set. $\dagger$ denotes using contrastive learning strategy in Section \ref{sec:mm}.
    }
    \label{tab:abl_repre}
\end{table}
}

\subsection{Ablation Study and Analysis}
\label{sec:ablation}

\noindent{\bf Importance of motion representations and feature space for matching.}
In Table \ref{tab:abl_repre}, we show the effectiveness of learning motion representations in a feature space for matching: 
(1) We represent tracklets and observations in the global coordinate by normalizing their 3D positions based on the ego-vehicle position, which is the scene-centric representation. We denote it as the global representation compared with our motion representation.
(2) Instead of matching in the feature space, we may associate tracklets and observations based on the distance between their output states (\eg, object position, heading angle, and size), which resembles the practice in the Kalman filter\cite{kalman1960kalman}.

In Table \ref{tab:abl_repre}, we observe from (a) $\rightarrow$ (c) and (b) $\rightarrow$ (d) that our motion representations are aware of object movements and thus help model training to achieve better performance.
Furthermore, from (a) $\rightarrow$ (b) and (c) $\rightarrow$ (d), we validate that matching in the feature space mitigates the potential noises in object states, which is important for monocular 3D MOT since the observations from the visual detector can be inaccurate. In addition, benefiting from matching in the feature space, (e) shows that the proposed contrastive loss learns more robust representations to further boost performance.

\setlength{\tabcolsep}{0.01\linewidth}{
\begin{table}[t]
    \footnotesize
    \centering
        
    \begin{tabular}{clccc}  
    
        \toprule & Ablation & AMOTA$\uparrow$ & AMOTP$\downarrow$ & MOTA$\uparrow$  \\ 
        \midrule
        (a) & Baseline \Bstrut & 27.1 & 1.465 & 23.4  \\ \hline%
        (b) & w/o Temporal encoder \Tstrut & 29.5 & 1.447 & 25.5 \\ %
        (c) & w/o Spatial encoder & 30.1 & 1.435 & 26.0 \\
        (d) & w/o Global positional feature & 30.7& 1.436& 26.9  \\ %
        (e) & Motion Transformer$\rightarrow$LSTM \Bstrut & 29.7 & 1.440 & 26.3 \\ \hline
        (f) & Full model \Tstrut & \textbf{31.1} & \textbf{1.432} & \textbf{27.1}\\ 
        \bottomrule
    \end{tabular}
    \vspace{-2.2mm}
    \caption{\textbf{Analysis of different components in the proposed motion transformer} using the nuScenes validation set.
    See Section \ref{sec:ablation} for details.
    }
    \label{tab:abl_arch}
\end{table}
}

\smallskip{\noindent{\bf Effectiveness of each component in motion transformer.}}
In Table \ref{tab:abl_arch}, we further investigate the effectiveness of each design in our motion transformer: mainly including the temporal encoder, spatial encoder, and the global positional feature in the spatial encoder. 

We show that each proposed module brings performance improvement. 
First, the baseline (a), without the whole proposed motion transformer, achieves undesirable performance (27.1 in AMOTA). 
In addition, comparing (b) with the full model (f), temporal learning provides the most improvement (+1.6 in AMOTA) since historical cues can help the model capture motion information. 
We also show the effect of using the spatial encoder and the global positional feature in the spatial encoder.
Results comparing (c)(d) to (f) show the importance of capturing the spatial interaction between different tracklets, as well as the awareness of 3D location to model the spatial interaction. 
Finally, we replace the proposed transformer architecture with the classical LSTM model (e) to show the effectiveness of the spatial-temporal modeling from our motion transformer.

\begin{figure*}[th]
\centering
\includegraphics[width=0.99\textwidth]{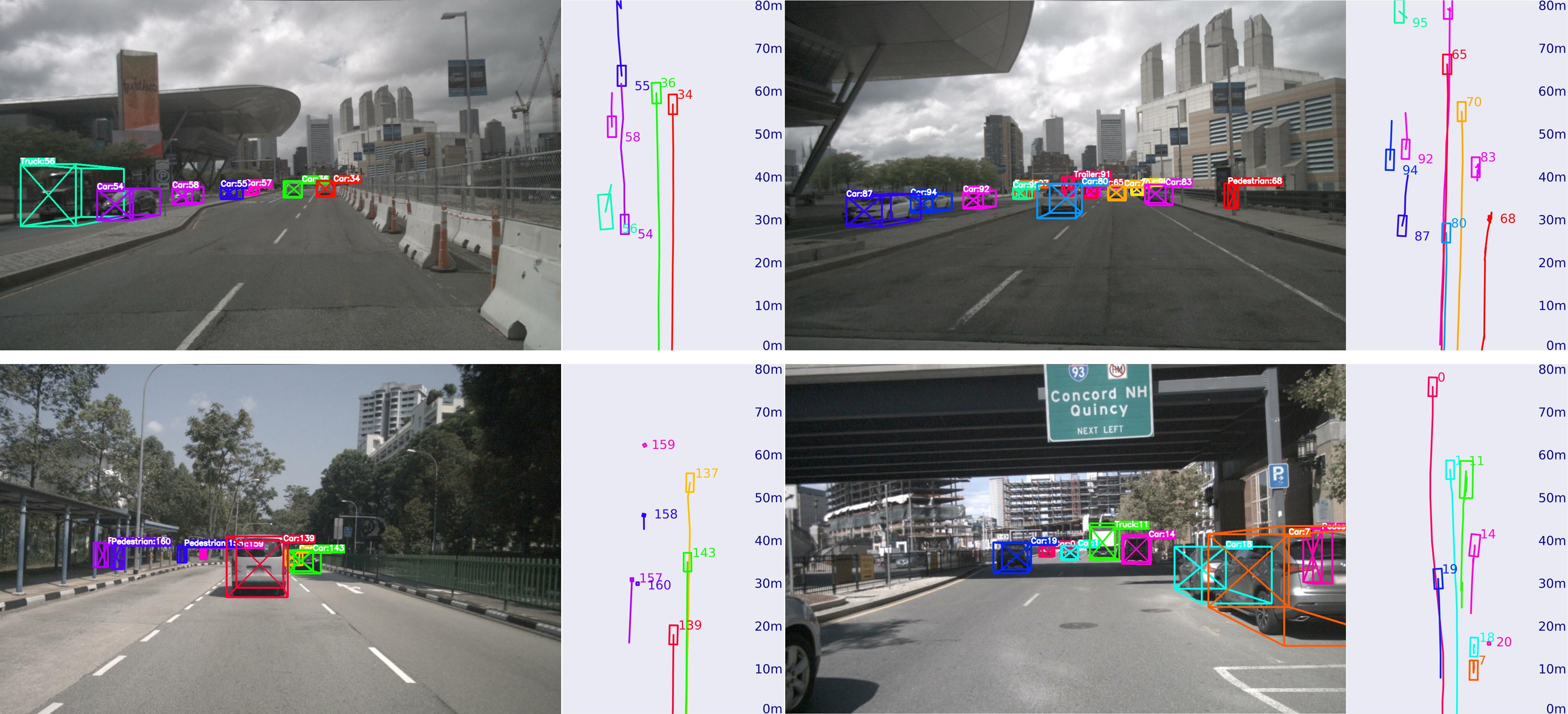}
\caption{\textbf{Qualitative visualization on the nuScenes validation set.} 
We provide some examples of tracking results on the image view for the current frame (left) and the trajectories in the bird's eye view (right) for 15 historical frames.
We utilize different colors and numbers to represent the different objects' identities. Best viewed in color and zoomed in.
}
\label{fig:nus_vis}
\end{figure*}

\begin{figure}[t]
    \centering
    \includegraphics[width=0.98\columnwidth]{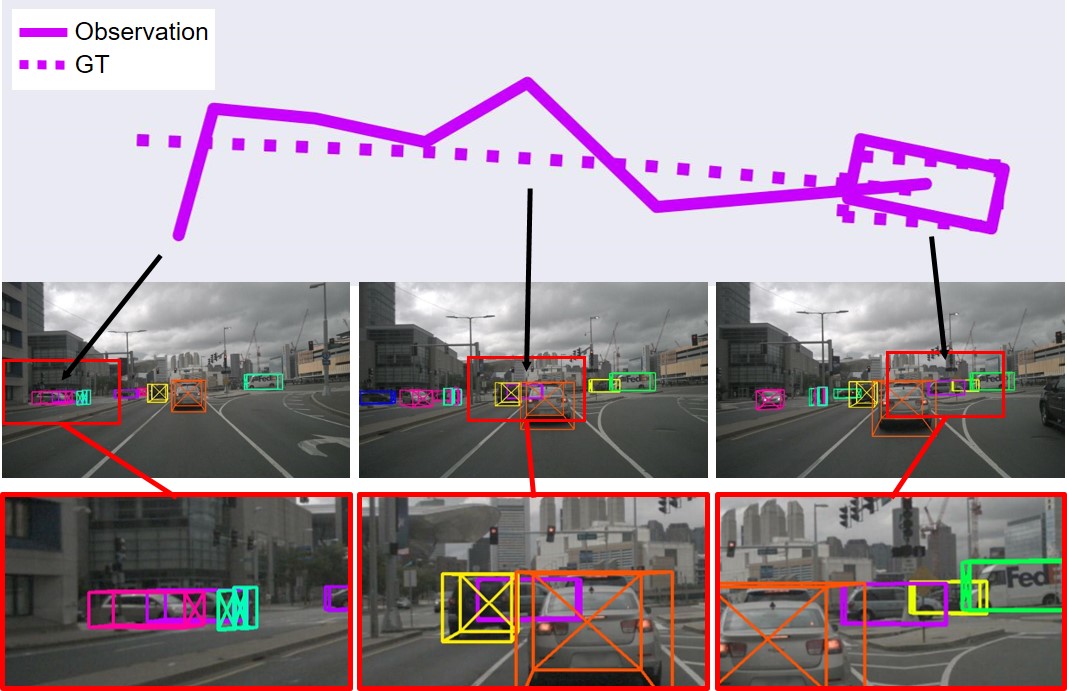}
    \caption{\textbf{Example results of handling inaccurate 3D observations.} 
    The noisy 3D detection results (\ie, the \textcolor{plotmagenta}{magenta} boxes) are often caused by occlusion and inaccurate predictions. The solid and dotted lines denote the observed and ground truth trajectories on the bird’s eye view plane. Our motion tracker is able to track objects even when the observations are not accurate enough.}
    \label{fig:robust_exp}
\end{figure}

\setlength{\tabcolsep}{0.04\linewidth}{
\begin{table}[t]
    \footnotesize
    \centering
    \begin{tabular}{l @{\hspace{10pt}} c @{\hspace{14pt}} c @{\hspace{8pt}} c}
        \toprule Detector & Method & AMOTA$\uparrow$ & AMOTP$\downarrow$\\ 
        \midrule
        \multicolumn{1}{l}{\multirow{3}{*}{FCOS3D 
        \cite{wang2021fcos3d}}} & KF3D     & 23.4 & 1.502    \\ %
         & LSTM  & 23.8 & 1.500    \\ %
        & \cellcolor{LightCyan}Ours   & \cellcolor{LightCyan}\textbf{26.0}  & \cellcolor{LightCyan}\textbf{1.447}  \\ 
        \midrule  %
        \multicolumn{1}{l}{\multirow{3}{*}{EPro-PnP \cite{epropnp}}} &KF3D  & 25.8 & 1.482   \\ %
        &LSTM & 27.0 & 1.470   \\ %
        &\cellcolor{LightCyan}Ours    & \cellcolor{LightCyan}\textbf{29.7} & \cellcolor{LightCyan}\textbf{1.418}    \\ %
        \bottomrule %
    \end{tabular}
    \vspace{-2mm}
    \caption{\textbf{Analysis of different motion modules} on the nuScenes validation set. We evaluate performance with different detectors using the same model checkpoint of trackers without re-training.
    }
    \label{tab:abl_motion}
\end{table}
}

\setlength{\tabcolsep}{0.005\linewidth}{
\begin{table}[!t]
    \centering
    \footnotesize
    \begin{tabular}{lccccc}  
        \toprule
        Method &  AMOTA$\uparrow$  & AMOTP$\downarrow$  & MOTA$\uparrow$ & MOTP$\downarrow$ & MOTAR$\uparrow$ \\ \midrule 
        MUTR3D\cite{zhang2022mutr3d}$\dagger$ & 27.0 & 1.494  & 24.5 & 0.709 & 64.3 \\ 
        CC-3DT\cite{cc3dt}$\diamond$ & 41.0 & 1.274  & 35.7 & \textbf{0.676} & 69.0 \\ 
        \midrule 
        MoMA-M3T (Ours)$\dagger$ & 36.0 & 1.349 & 31.1 & 0.700 & 68.4 \\ 
        MoMA-M3T (Ours)$\dagger^*$ & 41.5 & 1.278 & \textbf{36.8} & 0.701 & 71.0  \\ 
        MoMA-M3T (Ours)$\diamond$ & \textbf{42.5} & \textbf{1.240}  & 36.1 & 0.681 & \textbf{71.1} \\ 
        \bottomrule
    \end{tabular}
    \vspace{-2mm}
    \caption{\textbf{3D MOT performance on the nuScenes test set for the multi-camera tracking setting.} $\dagger$ and $\diamond$ denote using DETR3D~\cite{yue2021detr3d} and BEVFormer~\cite{li2022bevformer} as the detector with the ResNet101 \cite{He2015} backbone, respectively. $^*$ indicates the detector trained with a V2-99 backbone provided by \cite{park2021dd3d}.
    }
    \label{tab:nuscenes_cross}
\end{table}
}

\vspace{1mm}
\smallskip{\noindent{\bf Robustness analysis on monocular 3D object detectors.}}
To show the robustness of our tracker under various 3D detection outputs, we evaluate our motion tracker using different front-view-based monocular 3D object detectors, including FCOS3D \cite{wang2021fcos3d} and EPro-PnP \cite{epropnp}.
We perform tracking on these detectors using the same model without re-training to demonstrate the generalization ability of our motion tracker.

In Table \ref{tab:abl_motion}, we compare our motion modules with LSTM~\cite{hu2022monocular} and 3D Kalman filter~\cite{kalman1960kalman} that predict and match objects' state in the output space. We show that our MoMA-M3T achieves better performance across different detectors.
One of the main reasons is that matching in the feature space can perform more robustly than that in the output space, in which our method is less sensitive to noises from outputs of various 3D object detectors.

\vspace{1mm}
\smallskip\noindent{\bf Multi-camera 3D MOT on nuScenes.}
While our method focuses on the monocular setting, it is also applicable to multi-camera detection systems \cite{li2022bevformer, yue2021detr3d} that simultaneously recognize objects for all cameras, which can boost performance by filtering duplicate detections and benefiting tracking across cameras. We conduct experiments under the multi-camera tracking setting and present the tracking results on the nuScenes test dataset in Table \ref{tab:nuscenes_cross}.
We show that our MoMA-M3T obtains significant improvements of +9.0 in AMOTA compared to MUTR3D~\cite{zhang2022mutr3d} with the same detector (\ie, DETR3D~\cite{yue2021detr3d}), which indicates the effectiveness of our approach. 
Moreover, compared with CC-3DT\cite{cc3dt} using the same detector (\ie, BEVFormer\cite{li2022bevformer}), our approach obtains +1.5 in AMOTA.
This validates the effectiveness of adopting our method in various settings.

\subsection{Qualitative Results}
We show qualitative examples on the nuScenes validation set in Figure~\ref{fig:nus_vis} to illustrate that our motion tracker can track objects across various scenarios. 
Also, we provide a representative example in Figure~\ref{fig:robust_exp} to show that our motion-aware tracker can track objects well, even under inaccurate observations caused by occlusion and inaccurate depth estimation from the monocular 3D object detector.
More qualitative results are included in the supplementary material.

\section{Conclusions}
In this paper, we present MoMA-M3T, a motion-aware matching strategy for monocular 3D MOT. We represent the motion information for tracklets with their relative movements, followed by a motion transformer to model the motion cues from a spatio-temporal perspective. Consequently, a motion-aware matching module is applied to match tracklets and current observations based on their motion features. 
Extensive experiments on the nuScenes and KITTI datasets demonstrate that MoMA-M3T achieves state-of-the-art performance and is compatible to integrate with existing monocular 3D object detectors without the need of finetuning our tracker.


{\small
\bibliographystyle{ieee_fullname}
\bibliography{egbib}
}

\clearpage
\appendix
\noindent{\bf\Large Supplementary Material}
\section{Main Evaluation Metrics}
The nuScenes dataset evaluates the 3D MOT performance mainly by utilizing AMOTA, which is built upon the sAMOTA (scaled AMOTA) metric~\cite{Weng2020_AB3DMOT} to deal with the problem of MOTA~\cite{bernardin2008evalmot} that may tend to filter low-confidence detections because of the potential of causing false-positive results.
The AMOTA is defined as MOTA~\cite{bernardin2008evalmot} over $n$ recall thresholds:
\begin{align} \label{eq:disc}
     & {\rm AMOTA} =  \frac{1}{n} \sum_{r \in \{\frac{1}{n}, \frac{2}{n},\ldots,1\}} {\rm MOTAR}, \nonumber \\
     & {\rm MOTAR} = \max(0, 1- \frac{{\rm IDS}_r+{\rm FP}_r + {\rm FN}_r - (1-r)*{\rm GT}}{r*{\rm GT}}),
\end{align}
where ${\rm IDS}_r$, ${\rm FP}_r$, ${\rm FN}_r$ denote the number of identity switches, false positives, and false negatives calculated at the certain recall $r$, and $\rm{GT}$ is the number of ground truth.

\section{More Experimental Results}
\noindent{\textbf{Analysis of different frame lengths ($T$).}} 
In Table \ref{tab:abl_framelength}, we investigate the effect of different frame lengths utilized in our motion transformer. 
It is worth noting that we apply the global representation instead of the motion representation for the case of the single frame ($T=1$).
We find that, as the frame number becomes larger, the performance is improved gradually, especially when using more than 4 frames. In the main paper, we use $T=6$ as our final setting.

\smallskip\noindent{\textbf{Effectiveness of time positional encoding in motion transformer.}} 
In Table \ref{tab:abl_tpe}, we show that using a learnable time positional encoding for the proposed motion transformer improves the performance since it makes the model aware of motion cues at different timestamps. 

\smallskip\noindent{\textbf{3D MOT results on the nuScenes validation set.}} 
Table \ref{tab:nuscenes_val_single} and Table \ref{tab:nuscenes_val_cross} present the 3D tracking performance on the nuScenes validation set for single-camera and multi-camera settings. It shows that our MoMA-M3T achieves better results than existing methods on both tracking settings, which validates the effectiveness of our approach.

\setlength{\tabcolsep}{0.03\linewidth}{
\begin{table}[h]
    \footnotesize
    \centering
    \begin{tabular}{l @{\hspace{6pt}} c @{\hspace{9pt}} c @{\hspace{6pt}} c @{\hspace{6pt}} c}
        \toprule
        & Frame Number & AMOTA$\uparrow$ & AMOTP$\downarrow$ & MOTA$\uparrow$   \\ 
        \midrule
        (a) &   1  & 29.5 & 1.447 & 25.5 \\ %
        (b) &   2  & 30.2 & 1.441 & 26.2 \\  %
        (c) &   4  & 30.9 & 1.436 & \textbf{27.1} \\ %
        (d) &   6  & \textbf{31.1} & \textbf{1.432} & \textbf{27.1} \\ %
        \bottomrule
    \end{tabular}
    \vspace{-2mm}
    \caption{\textbf{Analysis of different frame lengths for our motion transformer} on the nuScenes validation set. We use the global representation to deal with the single frame observation.
    }
    \label{tab:abl_framelength}
\end{table}
}
\setlength{\tabcolsep}{0.03\linewidth}{
\begin{table}[ht]
    \footnotesize
    \centering

    \begin{tabular}{l  @{\hspace{9pt}} c @{\hspace{6pt}} c @{\hspace{6pt}} c}
        \toprule
        Setting & AMOTA$\uparrow$ & AMOTP$\downarrow$    & MOTA$\uparrow$\\ 
        \midrule
        w/o Time Positional Enc. & 30.7 & 1.435 & 26.4   \\ %
        w/ Time Positional Enc.   & \textbf{31.1} & \textbf{1.432}  & \textbf{27.1}   \\ %
        \bottomrule
    \end{tabular}
    \vspace{-2mm}
    \caption{\textbf{Effectiveness of time positional encoding} in motion transformer on the nuScenes validation set.
    }
    \label{tab:abl_tpe}
\end{table}
}
\setlength{\tabcolsep}{0.019\linewidth}{
\begin{table}[!t]
    \centering
    \footnotesize
    \begin{tabular}{@{}l@{\ \ \ }c@{\ \ }c@{\ \ }c@{\ \ }c@{\ \ }c@{}}
        \toprule
        Method &  AMOTA$\uparrow$  & AMOTP$\downarrow$  & RECALL$\uparrow$ & MOTA$\uparrow$ & MOTP$\downarrow$  \\ \midrule 
        CenterTrack \cite{zhou2020tracking}           & 6.8 & 1.54  & 0.23 & 6.1 & -  \\ 
        TraDeS \cite{Wu2021TraDeS}                  & 11.8 & 1.48 & 0.23 & -  & -  \\ 
        PermaTrack \cite{tokmakov2021learning}             & 10.9 & -    &   -  &  8.1 & -   \\
        DEFT \cite{Chaabane2021deft}                   & 20.9 & -    & -  & 17.8 & - \\ 
        Time3D \cite{li2022time3d}                  & 26.0 & \textbf{1.38}  & - & 20.7 & 0.82     \\ 
        QD-3DT \cite{hu2022monocular}          & 24.2 & 1.518 & 0.399 & 21.8 & 0.81 \\ 
        \midrule 
        MoMA-M3T (Ours) & \textbf{31.1} & 1.432 & \textbf{0.468} & \textbf{27.1} & \textbf{0.766} \\ 
        \bottomrule
    \end{tabular}
    \vspace{-2mm}
    \caption{\textbf{3D MOT performance on the nuScenes validation set for the single-camera tracking setting.} We use $\textbf{bold}$ to highlight the best results.}
    \label{tab:nuscenes_val_single}
\end{table}
}

\setlength{\tabcolsep}{0.015\linewidth}{
\begin{table}[!t]
    \centering
    \footnotesize
    \begin{tabular}{@{}l@{\ \ \ }c@{\ \ }c@{\ \ }c@{\ \ }c@{\ \ }c@{}}
        \toprule
        Method &  AMOTA$\uparrow$  & AMOTP$\downarrow$  & RECALL$\uparrow$ & MOTA$\uparrow$ & MOTP$\downarrow$  \\ \midrule
        MUTR3D \cite{zhang2022mutr3d}$\dagger$             & 29.4 & 1.498 & 0.427 & 26.7 & 0.799 \\
        CC-3DT \cite{cc3dt}$\diamond$ & 42.9 & 1.257 & 0.538 & 35.7 & -\\
        \midrule 
        MoMA-M3T (Ours)$\dagger$  & 36.2 & 1.369 & 0.484 & 31.2 & 0.794 \\ 
        MoMA-M3T (Ours)$\diamond$ & \textbf{44.8} & \textbf{1.225} & \textbf{0.550} & \textbf{38.8} & \textbf{0.714} \\ 
        \bottomrule
    \end{tabular}
    \vspace{-2mm}
    \caption{\textbf{3D MOT performance on the nuScenes validation set for the multi-camera tracking setting.} $\dagger$ and $\diamond$ denote using DETR3D~\cite{yue2021detr3d} and BEVFormer~\cite{li2022bevformer} as the detector with the ResNet101 \cite{He2015} backbone, respectively.
    }
    \label{tab:nuscenes_val_cross}
\end{table}
}

\section{Qualitative Visualization}

\begin{figure*}[ht]
\centering
\includegraphics[width=0.99\textwidth]{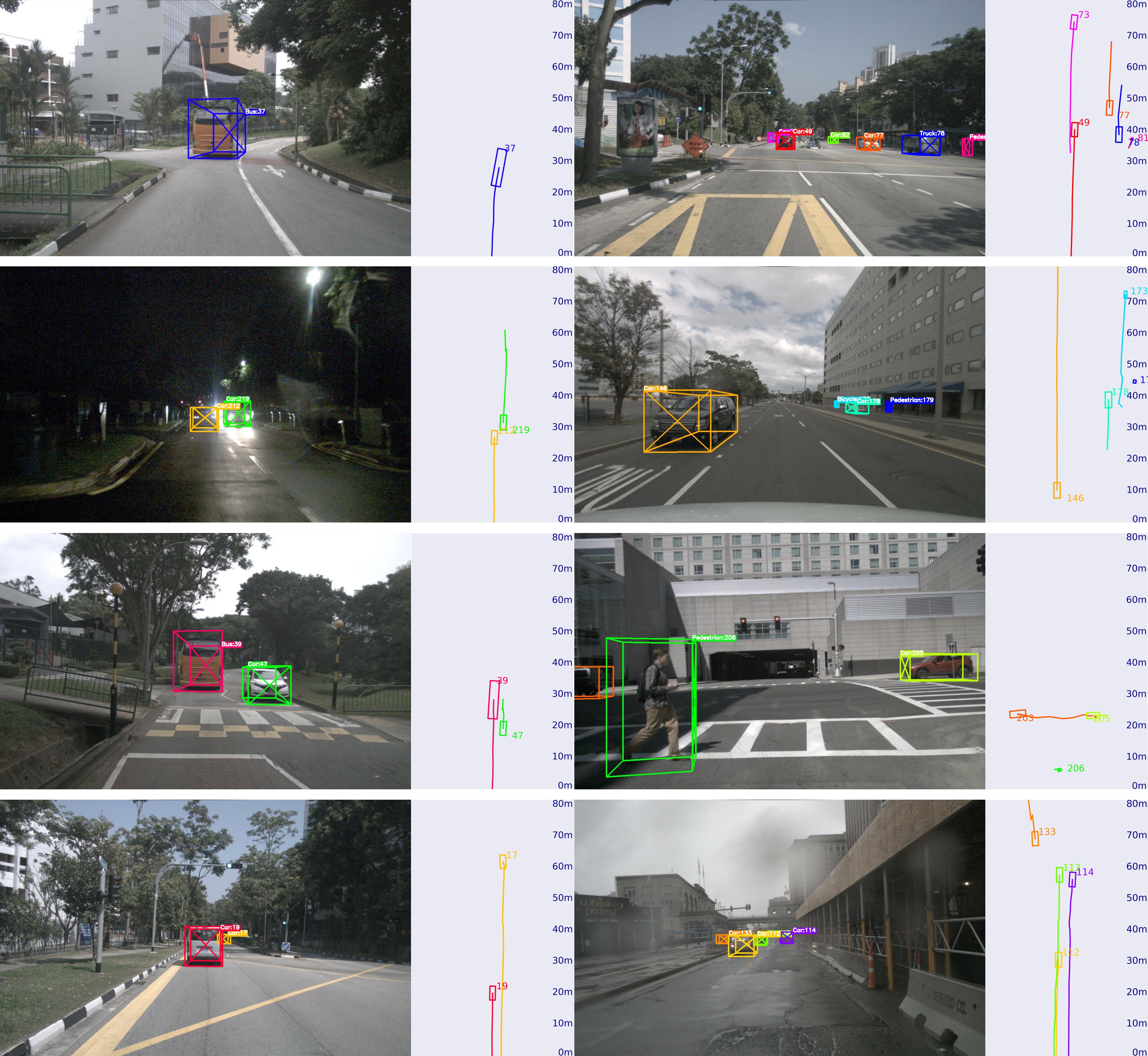}
\caption{\textbf{Qualitative results on the nuScenes validation set.} We plot the tracking results of our MoMA-M3T based on the image view (left) and the bird's eye view (right) with the 15 historical frames on the BEV plane, in which different colors denote different tracklets.
}
\label{fig:supp_nus_vis}
\vspace{-6pt}
\end{figure*}

\begin{figure*}[ht]
\centering
\includegraphics[width=0.99\textwidth]{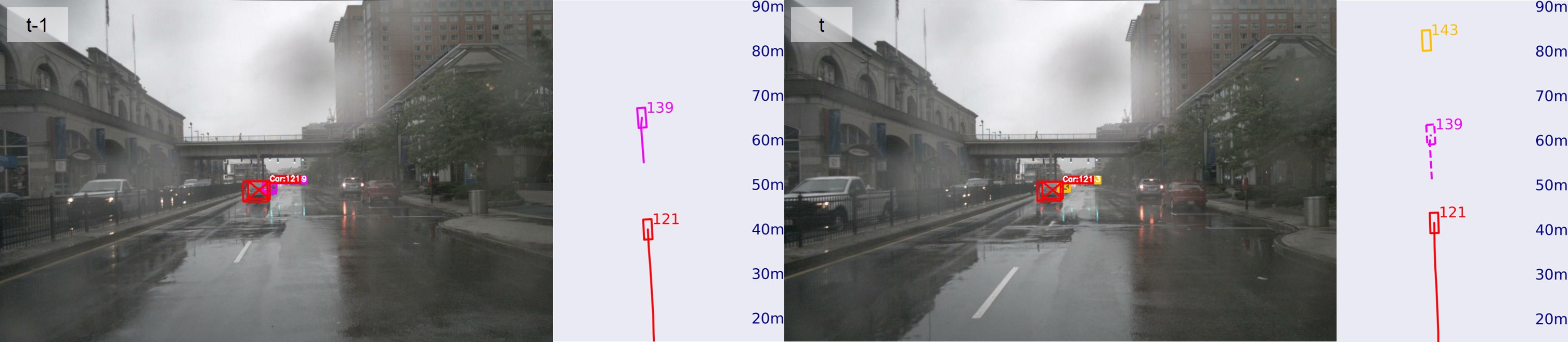}
\caption{\textbf{Representative failure case.} 
The solid lines and boxes denote the tracking results for timestamps $t-1$ and $t$, while the dotted one is the mismatched tracklet (\text{\#139}) in the coordinate at $t$ timestamp. 
The failure case is caused by the inaccurate observation (yellow box at frame $t$) from the monocular 3D object detector. 
The tracker cannot associate the detection with any tracklet (\eg, \text{\#139}), thus generating a new identity for it (\text{\#143}). Note that, we only plot a few bounding boxes for better illustrations in this example.
}
\label{fig:supp_nus_fail}
\vspace{-6pt}
\end{figure*}

\noindent{\textbf{More visualization results.}}
In Figure \ref{fig:supp_nus_vis}, we show example visualization results on the nuScence validation set. It can be observed that our method can track different types of objects across various scenarios.

\smallskip{\noindent{\textbf{Failure case.}} 
We provide a representative failure case in Figure \ref{fig:supp_nus_fail}. 
Due to the inaccurate object depth estimation for the yellow box (see the right figure), the tracker cannot associate it with the existing tracklet ($\text{\#139}$) since their position distance is too far from each other (more than 10 meters). It thus generates a new identity ($\text{\#143}$) for the detection.

}

\end{document}